\documentclass{bmvc2k}
\usepackage{multirow}
\usepackage{footnote}
\usepackage{float}
\usepackage{threeparttable}

\title{Spatio-Temporal Action Detection with Cascade Proposal and Location Anticipation}

\addauthor{Zhenheng Yang}{zhenheny@usc.edu}{1}
\addauthor{Jiyang Gao}{jiyangga@usc.edu}{1}
\addauthor{Ram Nevatia}{nevatia@usc.edu}{2}

\addinstitution{
Institute for Robotics and Intelligent Systems\\
 University of Southern California\\
 Los Angeles, CA, USA
}

\runninghead{Zhenheng Yang, Jiyang Gao, Ram Nevatia}{Spatio-Temporal Action Detection with Cascade Proposal and Location Anticipation}

\begin{document}

\maketitle

\begin{abstract}
In this work, we address the problem of spatio-temporal action detection in temporally untrimmed videos. It is an important and challenging task as finding accurate human actions in both temporal and spatial space is important for analyzing large-scale video data. To tackle this problem, we propose a cascade proposal and location anticipation (CPLA) model for frame-level action detection. There are several salient points of our model: (1) a cascade region proposal network (casRPN) is adopted for action proposal generation and shows better localization accuracy compared with single region proposal network (RPN); (2) action spatio-temporal consistencies are exploited via a location anticipation network (LAN) and thus frame-level action detection is not conducted independently. Frame-level detections are then linked by solving an linking score maximization problem, and temporally trimmed into spatio-temporal action tubes. We demonstrate the effectiveness of our model on the challenging UCF101 and LIRIS-HARL datasets, both achieving state-of-the-art performance.

\end{abstract}

\section{Introduction}
\label{sec:intro}

We aim to address the problem of action detection with spatio-temporal localization: given an untrimmed video, the goal is to detect and classify every action occurrence in both spatial and temporal extent. Advances in convolutional neural network (CNN) have triggered improvements in video action recognition \cite{simonyan2014two,gkioxari2015contextual,gan2016webly}. Compared with action recognition, spatio-temporal action detection is more challenging due to arbitrary action volume shape and large spatio-temporal search space.

There has been previous work in spatio-temporal action detection. Recent deep learning based approaches \cite{gkioxari2015finding,weinzaepfel2015learning,suman16bmvc,peng2016multi} first detect actions independently on the frame-level which are then either linked or tracked to form a final spatio-temporal action detection result. These methods use both appearance and motion features but process them separately and then fuse the detection scores. However, an action occurrence extends over a period and there should be consistency within the movement of the action regions at different temporal points.
For example in the action "diving", the swinging up of arms often indicates a lunge and jump in around a second and head-down diving in another second. Thus the location of the action in one or two seconds will be lower than the current location. With such consistencies, the relationship of action spatial localization in time can be leveraged and modeled.


\begin{figure}
\centering
\includegraphics[width=1.0\textwidth]{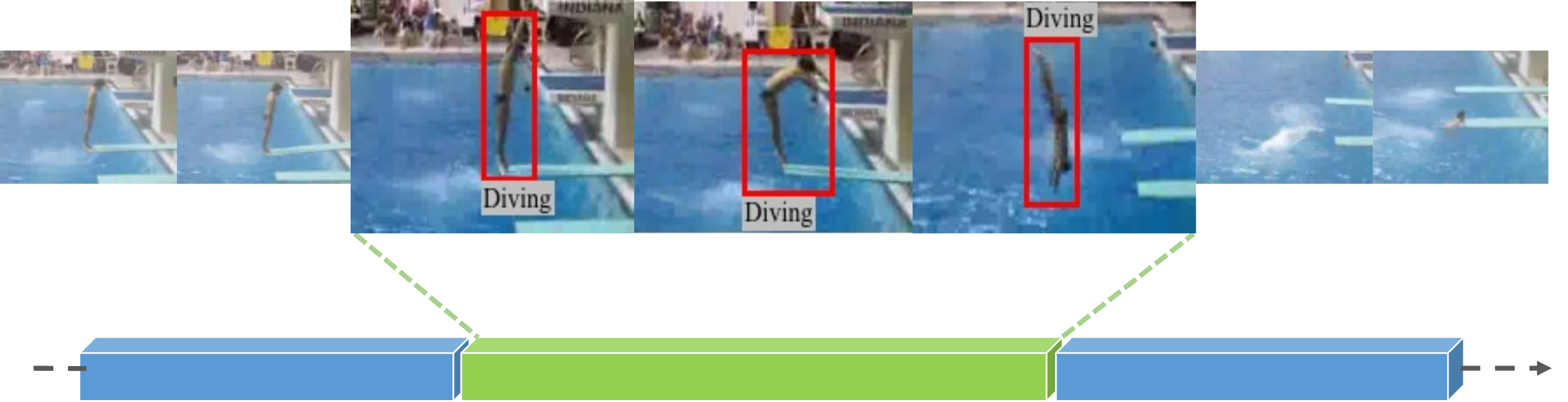}
\caption{Spatio-temporal action detection in an untrimmed video.}
\label{fig:prob_illu}
\end{figure}


We propose to use Cascade Proposal and Location Anticipation (\textit{CPLA}) for Action Detection in videos. Specifically, our approach consists of spatial detection and temporal linking. The frame-level spatial detection model is composed of three parts: cascade RPN (\textit{casRPN}) as action proposal generator, a network similar to fast R-CNN as the detection network and a location anticipation network (\textit{LAN}). LAN is designed for inferring the movement trend of action occurrences between two frames, $t-K$ and $t$ where $t>K$, $K$ is the \textit{anticipation gap}. LAN takes detected bounding boxes from detection network output on frame $t-K$ and then infers the corresponding boxes on frame $t$. The anticipated bounding boxes serve as additional proposals for the detection network on frame $t$. To exploit both appearance and motion cues, two-stream networks are implemented for all casRPN, detection network and LAN.

We implement the casRPN as a two stage cascade of RPNs, based on our observation that original RPN suffers from low recall at high intersection-over-union (IoU). The casRPN takes images (RGB images or optical flow images) as inputs and outputs spatial action proposals. Detection network takes proposals as input and further classifies and regresses to detection results. Spatial action detection results are temporally linked and trimmed to produce a spatio-temporal action tube, similar to \cite{suman16bmvc,peng2016multi}.

The proposed approach is assessed on UCF101 \cite{soomro2012ucf101} 24 classes (UCF101-24) and LIRIS-HARL \cite{wolf2014evaluation} datasets for proposal performance and spatio-temporal action detection performance. casRPN outperforms other proposal methods \cite{uijlings2013selective,zitnick2014edge,ren2015faster} by a large margin on both datasets, especially in the high IoU range. For action detection performance, CPLA achieves state-of-the-art performance on both datasets. For example, an mAP of 73.54\% is achieved at standard spatio-temporal IoU of 0.2 on UCF101-24, an improvement of 0.68\% from 72.86\% , the current state-of-the-art method \cite{peng2016multi}.

In summary, our contributions are three-fold:

(1) We propose a location anticipation network (LAN) for action detection in videos that exploits the spatio-temporal consistency of action locations.

(2) We propose a cascade of region proposal network (casRPN) for action proposal generation which achieves better localization accuracy.

(3) We comprehensively evaluate different variants of CPLA on UCF101-24 and LIRIS-HARL datasets and CPLA achieves state-of-the-art performance.

\section{Related work}

Inspired by the advances in image classification and object detection on images, the deep learning architectures have been increasingly applied to action recognition, temporal action detection, spatial action detection and spatio-temporal action detection in videos.

\textbf{R-CNN for Object detection.} R-CNN \cite{girshick14CVPR} has achieved a significant success in object detection in static images. This approach first extracts proposals from images with selective search \cite{uijlings2013selective} algorithm and then feeds the rescaled proposals into a standard CNN network for feature extraction. A support vector machine (SVM) is then trained on these features and classifies each proposal into one of object categories or background. There are a sequence of works improving R-CNN \cite{girshick14CVPR}. SPP-net \cite{he2014spatial} implements a spatial pyramid pooling strategy to remove the limitation of fixed input size. Fast R-CNN \cite{girshick2015fast} accelerates R-CNN by introducing a ROI pooling layer and improve the accuracy by implementing bounding box classification and regression simultaneously. Faster R-CNN \cite{ren2015faster} further improves the speed and performance by replacing proposal generation algorithm with a region proposal network (RPN).

\textbf{Action spatial detection and temporal detection.} There have been considerable works on spatial action detection in trimmed videos and temporal detection in untrimmed videos. On spatial action detection, Lu \textit{et al.} \cite{lu2015human} propose to use both motion saliency and human saliency to extract supervoxels and apply a hierarchical Markov Random Field (MRF) model for merging them into a segmentation result. Soomro \textit{et al.} \cite{soomro2015action} further improve the performance by incorporating spatio-temporal contextual information into the displacements between supervoxels. Wang \textit{et al.} \cite{wang2016actionness} first apply a two-stream fully convolutional network to generate an action score map (called ``actionness map''). Then action proposals and detections are extracted from the actionness map. 

For action temporal detection, sliding window based approaches have been extensively explored \cite{gaidon2013temporal,tian2013spatiotemporal,wang2014video}. Bargi \textit{et al.} \cite{bargi2012online} apply an online HDP-HMM model for jointly segmenting and classifying actions, with new action classes to be discovered as they occur. Ma \textit{et al.} \cite{ma2016learning} address the problem by applying a LSTM network to generate detection windows based on frame-wise prediction score. Singh \textit{et al.} \cite{singh2016multi} extends two-stream networks to multi-stream LSTM networks. S-CNN \cite{scnn_shou_wang_chang_cvpr16} propose a two-stage action detection framework: first generate temporal action proposals and then score each proposal with a trained detection network.

\textbf{Spatio-temporal action detection.} Although there have been a lot of efforts on both spatial action detection and temporal action detection, only a handful of efforts have been devoted to the joint problem of localizing and classifying action occurrences in temporally untrimmed videos. Tian \textit{et al.} \cite{tian2013spatiotemporal} extend 2D deformable part model \cite{felzenszwalb2008discriminatively} to action detection in videos. Jain \textit{et al.} \cite{jain2014action} use super-voxels to find the action boundaries. More recent works leverage the power of deep learning networks. Gkioxari and Malik \cite{gkioxari2015finding} extract proposals on RGB with selective search algorithm, and then apply R-CNN network on both RGB and optical flow data for action detection per frame. The frame-level detections are linked via Viterbi algorithm. Wainzaepfel \textit{et al.} \cite{weinzaepfel2015learning} replace selective search with a better proposal generator, i.e. EdgeBoxes \cite{zitnick2014edge} and conduct tracking on some selected frame-level action detections. Mettes \textit{et al.} \cite{mettes2016spot} propose to use sparse points as supervision to generate proposals. The two most recent works \cite{suman16bmvc,peng2016multi} extend faster R-CNN in static images and train appearance and motion networks for frame-level action detection independently. The detections of two-stream networks are late fused and linked via Viterbi algorithm \cite{forney1973viterbi}. A temporal trimming is then applied to generate spatio-temporal action tubes.

\section{Methodology}
\label{methodology}
\begin{figure}
\centering
\includegraphics[width=1.0\textwidth]{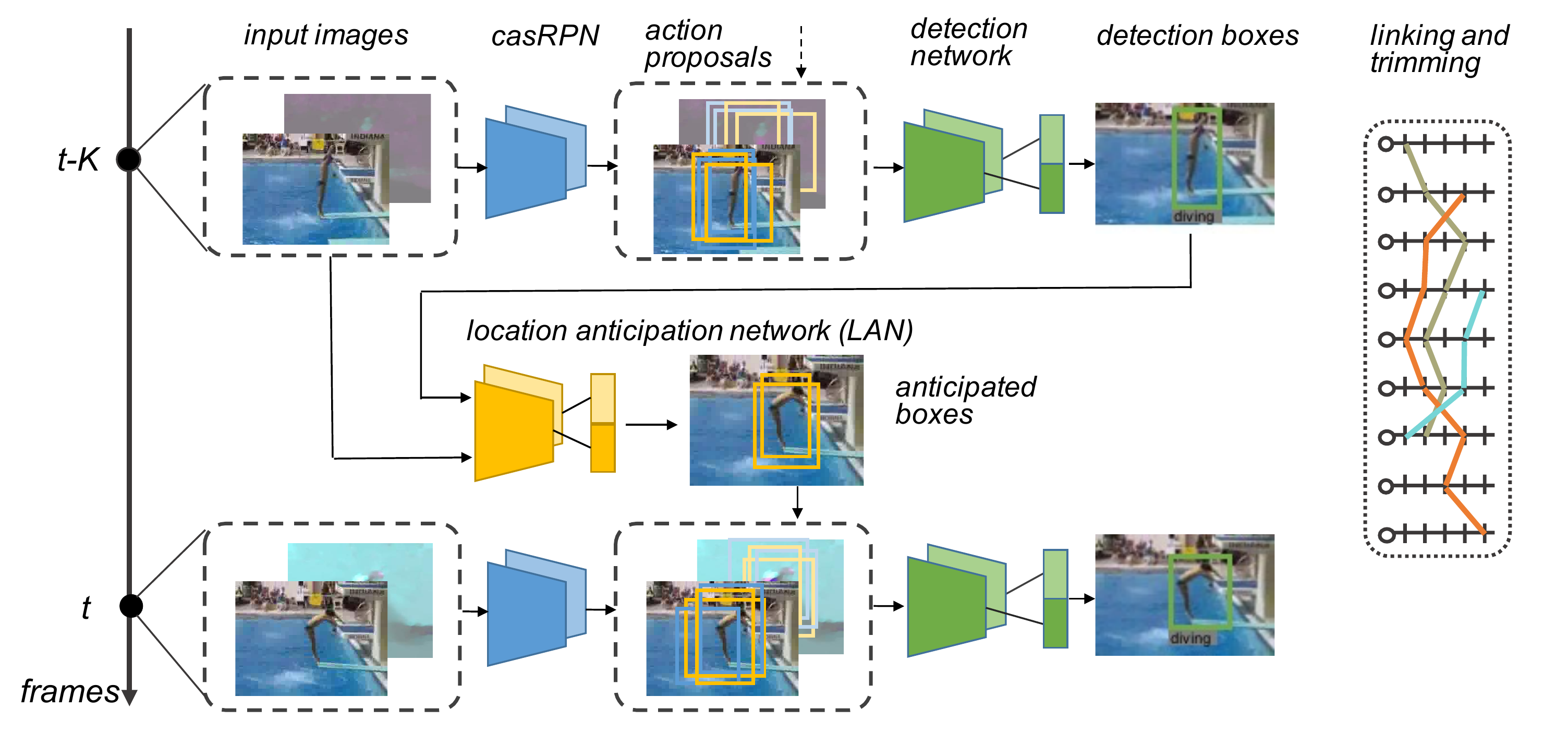}
\caption{Overview of cascade proposal and location anticipation (CPLA) model for action detection. Inputs are RGB and optical flow images, processed by casRPN and detection network. Location anticipation network (LAN) leverages the temporal consistency by inferring action region movement. The right most part represents detection linking and trimming. The horizontal and vertical lines represent different frames and detections on each frame. Three lines linking action detections across frames represent generated spatio-temporal action tubes.}
\label{fig:overview}
\end{figure}

As shown in Figure \ref{fig:overview}, a proposal network (casRPN), detection network and LAN are combined to generate frame-level action detections. The outputs are then linked and temporally trimmed to generate action tubes. In this section, the model architecture and training procedures are discussed in detail.

\subsection{Proposal Generation Network}
\label{casRPN}
\begin{figure}
\centering
\includegraphics[width=1.0\textwidth]{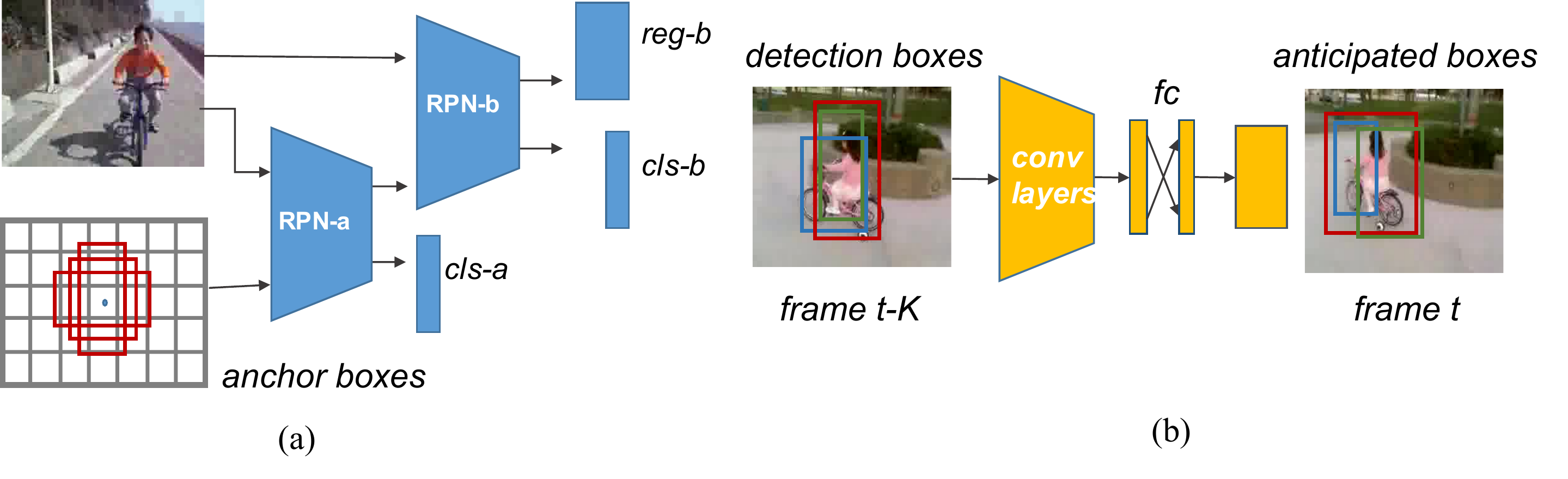}
\caption{Network architectures: (a) casRPN and (b) LAN.}
\label{fig:cas_rpn}
\end{figure}

We adopt a two-stage cascade of RPNs \cite{ren2015faster} that we call \textit{casRPN}. Similar to \cite{ren2015faster}, each stage is built on top of the last convolutional layer of the VGG-16 network \cite{simonyan2014very} followed by two sibling output layers: classification ($cls$) and regression ($reg$). To generate region proposals, the original RPN slides over the feature map output by the last convolutional layer and takes reference bounding boxes (called \textit{"anchor boxes"}) as input, outputting the objectness score and bounding box regression coordinates. For the first stage of casRPN (\textit{RPN-a}), we follow the anchor box generation process as in \cite{ren2015faster}. The proposals generated from RPN-a serve as the anchor boxes of the second RPN (\textit{RPN-b}) for scoring and another round of regression. Final proposal results are \textit{reg-b} and \textit{cls-b} generated from RPN-b. More details of the architecture are shown in Figure \ref{fig:cas_rpn} (a).

\textbf{Training.} The two stages of VGG-16 net are trained independently. A training protocol similar to \cite{ren2015faster} is followed: anchors with a high Intersection-over-Union (IoU) with the ground truth boxes (IoU > 0.7) are considered as positive samples, while those with low IoU (IoU < 0.3) as negative samples. Considering that in the action datasets (such as UCF101), there are fewer occurrences in one frame compared to those in the object detection datasets (such as Pascal VOC 2007 \cite{pascal-voc-2007}). 
To achieve network fast convergence, we ensure that in each mini-batch, the positive-negative sample ratio is between 0.8 and 1.2, and also that the mini-batch size is no larger than 128. The learning rate is set to be 0.0005. The Adam \cite{kingma2014adam} optimizer is used.  

\subsection{Detection network}
The detection network is built upon the last convolutional layer of VGG-16 network \cite{simonyan2014very} followed by a fusion of two stream features, two fully connected (\textit{fc}) layers and then two sibling layers as regression and classification. The detection network takes the images (RGB and optical flow images) and proposals as input and regresses the proposals to a new set of bounding boxes for each class. Each proposal leads to $C$ (number of classes) bounding boxes and corresponding classification scores. The network architecture is similar to that of fast R-CNN \cite{girshick2015fast} except that the two-stream features are concatenated at $conv5$ level.

\textbf{Training.} For detection network training, a variation of training procedure in \cite{ren2015faster} is implemented. Shaoqing \textit{et al.} \cite{ren2015faster} introduced a four-step `alternating training' strategy: RPN and detection network (fast R-CNN) are trained independently in the first two steps while in the 3rd and 4th step, the two networks are fine-tuned with shared convolutional weights. In our method, the casRPN and detection network are trained independently following the first two steps as we found the detection accuracy decreases when using shared convolutional weights. During training, the number of proposal bounding boxes and learning rate are set to be 2000 and 0.001. Adam optimizer \cite{kingma2014adam} is employed.

\subsection{Location Anticipation Network}
We use the location anticipation network (LAN) to predict the movement of action occurrences within an anticipation gap $K$ (frames). The input to LAN are RGB images and optical flow images and the action detection results on frame $t-K$. Regions of interest (ROIs) are extracted from input images and processed by each stream of LAN. The two stream features computed from RGB and optical flow images are then concatenated and the anticipated bounding boxes are generated from a regression layer applied on the fused features. The inferred action bounding boxes serve as additional proposals and fed into the detection network on frame $t$ 
The LAN is built upon the last convolutional layer of fast R-CNN \cite{girshick2015fast} network followed by the fusion of two stream features and $fc$ regression layers. The architecture of LAN is illustrated in Figure \ref{fig:cas_rpn} (b). 

\textbf{Two stream fusion.}
As the inputs to motion stream of the network are optical flow images, they already contain action movement information; we propose to leverage the coupling of appearance and motion cues to provide more prediction information by concatenating two-stream \textit{conv5} features before $fc$ layers in LAN.

\textbf{Training.} For LAN, the input contains the detection results on frame $t-K$ ($t>K$) and the target is the ground truth bounding boxes of frame $t$. Similar to the training protocol for casRPN \ref{casRPN}, bounding boxes in frame $t$ having high IoU (IoU > 0.7) with frame $t$ ground truth boxes are considered positive samples while low IoU (IoU < 0.3) are taken to be negative samples. The loss function for an image is defined as: \begin{equation}
L(t_i) = \frac{1}{N_{reg}}\sum_ip_i^*L_{reg}(t_i, t_i^*) 
\end{equation}
where, $i$ is the index of a detection bounding box, $t_i$ is a vector representing the parameterized bounding box coordinates, $t_i^*$ is the associated ground truth bounding box, $N_{reg}$ is the number of detection bounding boxes, the ground truth label term $p_i^*$ is 1 if the detection box is positive and 0 if it is negative. The ground truth label $L_{reg}(t_i, t_i^*)$ is a smooth $L_1$ loss function as in \cite{schmidt2007fast}. The mini-batch size and learning rate are set to be 300 and 0.001 separately and Adam optimizer \cite{kingma2014adam} is deployed for minimizing the loss function above.

\subsection{Detection linking and temporal trimming.} 

We employ detection linking using the Viterbi algorithm \cite{forney1973viterbi} and apply maximum subarray algorithm for temporal trimming as in \cite{peng2016multi}.

\textbf{Detection linking.} The linking score function $S_c(d_t, d_{t+1})$ of linking action detections of class $c$ in two consecutive frames is defined as below
\begin{equation}
\label{equ:energy}
S_c(d_t, d_{t+1}) = {(1-\beta)*(s_c(d_t) + s_c(d_{t+1}))+\beta*\Psi(d_t,d_{t+1})}
\end{equation}
in which, $s_c(d_i)$ is the class score of detection box at frame $i$, $\Psi(d_t, d_{t+1})$ is the IoU between two detection boxes, $\beta$ is a scalar weighting the relative importance of detection score and overlaps. The linking score is high for those links in which detection boxes score high for the action class $c$ and also overlap highly in consecutive frames. We find paths with maximum linking scores via Viterbi algorithm \cite{forney1973viterbi}. $\beta$ is empirically set to be 0.7.

\textbf{Temporal trimming.} In realistic videos, there is no guarantee that human actions occupy the whole span thus temporal trimming is necessary for spatio-temporal action localization. Inspired by \cite{an2009efficient}, we employ an optimal subarray algorithm similar to \cite{peng2016multi}. Given a video track $\Gamma$, we aim to find a subset starting from frame $s$ to frame $e$ within $\Gamma$. The optimal subset $\Gamma_{(s,e)}$ maximizes the objective:
\begin{equation}
\frac{1}{(e-s)}\sum_s^e S_c(d_t, d_{t+1})-\frac{(e-s)-\overline{L_c}}{\overline{L_c}}
\end{equation}

In this objective function, $s$ and $e$ are the indexes of starting and ending frame of the track subset. $S_c(d_t, d_{t+1})$ is the linking score as in \ref{equ:energy}. $\overline{L_c}$ is the average length (in frames) of action category $c$ in training data. The objective function aims to maximize the average linking scores between two frames in a track and to minimize the drift of generated track length from the average track length. 

\section{Evaluation}

We first introduce the datasets and evaluation metrics used in our experiments and then present a comprehensive evaluation of our methods.

\subsection{Datasets and metrics.}

\textbf{Datasets.} Two widely used datasets are selected for evaluating the performance of our action proposals and spatio-temporal action detection: (1) UCF101 \cite{soomro2012ucf101} 24 classes (UCF101-24) and (2) LIRIS-HARL \cite{wolf2014evaluation}.

(1) UCF101-24 is a subset of larger UCF101 action classification dataset. This subset contains 24 action classes and 3207 temporally-untrimmed videos, for which spatio-temporal ground truths are provided. As in \cite{suman16bmvc,peng2016multi}, we evaluate on the standard training and test split of this dataset.  (2) LIRIS-HARL dataset contains temporally-untrimmed 167 videos of 10 action classes. Similar to the UCF101-24 dataset, spatio-temporal annotations are provided. This dataset is more challenging, as there are more action co-occurrences in one frame, more interacting actions with humans/objects, and cases where relevant human actions take place among other irrelevant human motion.

\textbf{Evaluation metrics.}
The evaluation metrics in original papers \cite{soomro2012ucf101,wolf2014evaluation} are followed for separate dataset. Specifically, for assessing the quality of proposals, \textit{recall-vs-IoU} curve is employed. When measuring the action detection performance on UCF101-24, \textit{mean Average Precision} (\textit{mAP}) at different spatio-temporal IoU threshold of \{0.05,0.1,0.2,0.3,0.4,0.5\} is reported, using the evaluation code of \cite{suman16bmvc}. The official evaluation toolbox of LIRIS-HARL is used to measure the spatial precision and recall, along with temporal precision and recall, and finally integrating into a final score. Different metrics are reported: $Recall_{10}$, $Precision_{10}$, F1-$Score_{10}$, etc. More details of evaluation metrics are available in the original paper. The mAP metric performance as in UCF101-24 is also presented for better comparison. 

\subsection{Experiments}

Several experiments are conducted on UCF101-24 and LIRIS-HARL dataset for an comprehensive evaluation of CPLA approach: (1) casRPN is compared with other proposal methods for proposal quality assessment; (2) Two different anticipation strategies and non-anticipation model are compared. (3) Different anticipation gaps $K$ are explored and discussed. (4) Two fusion methods are explored and compared. (5) The CPLA model is compared with state-of-the-art methods on UCF101-24 and LIRIS-HARL datasets in terms of spatio-temporal action detection performance.

\begin{figure}
\centering
\includegraphics[width=1.0\textwidth]{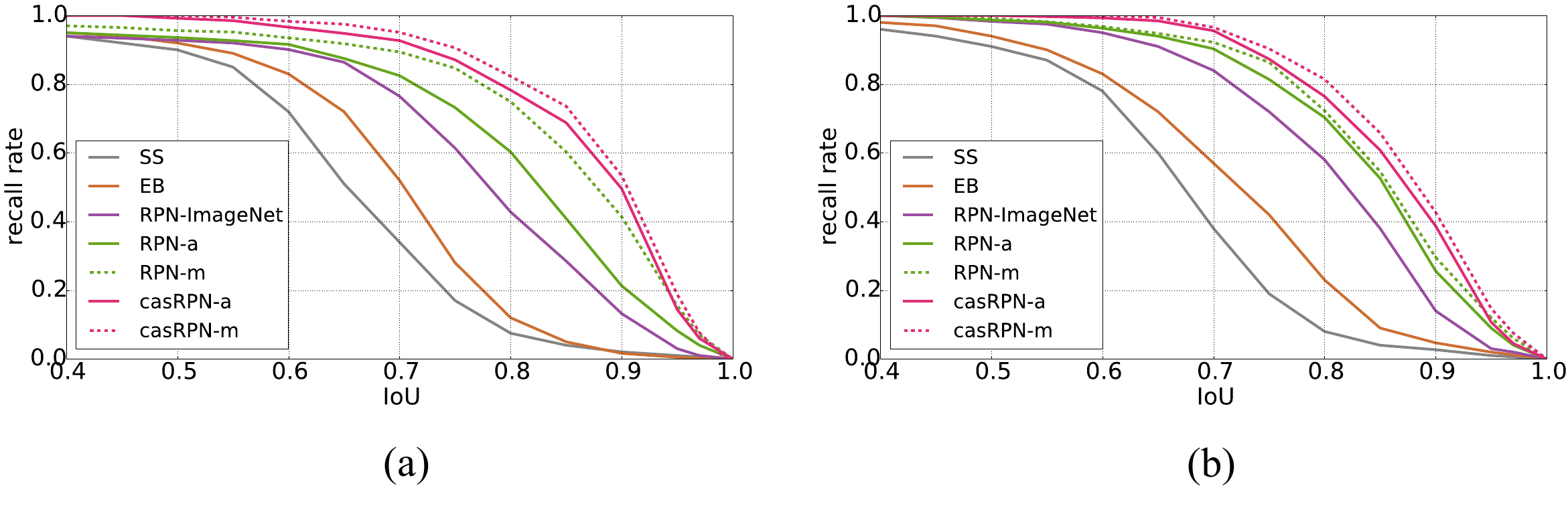}
\caption{Performance comparison of action proposal methods on (a) UCF101-24 split1 and (b) LIRIS-HARL.}
\label{fig:prop_perf}
\end{figure}

\textbf{Evaluation of casRPN performance.} The quality of casRPN proposals are compared with several other proposal methods on UCF101-24 split1. These methods include Selective Search (SS) \cite{uijlings2013selective}, EdgeBoxes(EB) \cite{zitnick2014edge}, RPN trained on ImageNet, RPN trained on UCF101-24 split1 (both appearance and motion models, RPN-a, RPN-m) and casRPN trained on UCF101-24 split1 (casRPN-a, casRPN-m). For Selective Search and EdgeBoxes, the default settings are implemented (2000 and 1000 proposals are extracted separately from one image). While for RPN based methods, top 300 proposals are picked. The recall-vs-IoU curves are plotted for evaluating proposal quality. Figure \ref{fig:prop_perf} shows that even with a relatively smaller number of proposals, RPN based proposals consistently exhibit much better recall performance compared to the two non deep-learning methods. casRPN outperforms other RPN methods by a large margin, especially at high IoU region, validating the effectiveness of two stage of regression in accurate localization.

\textbf{Anticipation strategy study.} We compare different anticipation strategies: (1) Original CPLA model. (2) The detection bounding boxes from frame $t-K$ are directly fed as proposals to detection network of frame $t$ (Non-motion CPLA). In this case, zero motion is assumed during the prediction gap $K$. (3) Detections on different frames are conducted independently and the proposals of each frame only come from the casRPN (Non-anticipation model). For all three different anticipation models, casRPN is employed as proposal generator and the performances are compared with mAP metric on the testing part of UCF101-24 split 1. As shown in Table \ref{predictive_perf}, the models that exploit the spatio-temporal consistency of actions locations benefit from the additional information compared with independent frame-level detection. At the standard threshold of $\delta = 0.2$, CPLA achieves mAP of 73.54\%. Comparing the two different anticipation strategies, we can see that a trained anticipation model (CPLA) outperforms the naive non-motion model (Non-motion CPLA) by 3.11\% at $\delta = 0.2$ when $K=8$. These comparisons indicate that anticipation model leveraging the consistency of action locations in temporal space helps boost the performance of spatio-temporal action detection. We also explored use of multiple frames within the anticipation gap and feeding their detection results as inputs to LAN. Uniformly sampling 2 frames from the anticipation gap shows 2.4\% mAP performance drop on UCF101-24 at $\delta=0.2$. This can be explained that doubling the input size, the number of parameters of $fc$ layer also double and thus LAN may be overfitting.

\textbf{Exploring different anticipation gaps.} Different choices of the anticipation gaps $K$ are also explored: $K$ is set to be \{2,8,16\} respectively to see how it affects the mAP performance. As shown in Table \ref{predictive_perf}, $K = 8$ shows the best result and outperforms the other anticipation gaps by at least 2.50\% at $\delta=0.2$. The performance decreases with too short ($K=2$) or too long ($K=16$) gaps. For too short anticipation gap, the network is predicting the movement between a very short duration of 0.06s. No obvious motion happens in this short time and thus it shows similar performance to zero motion model. With too long an anticipation gap, the performance drops. It can be explained by the observation that temporal cues are too noisy when looking at a relatively long time before the action occurs.

\begin{table}[!htbp]
\centering
\caption{Action detection performances under different anticipation strategies}
\label{predictive_perf}
\begin{tabular}{l|c|cccc}
\hline
\multicolumn{2}{l}{Spatio-temporal overlap threshold ($\delta$)}                    & 0.05  & 0.1   & 0.2   & 0.3   \\ \hline
\multicolumn{2}{l|}{Non-anticipation model}                              & 77.32 & 76.71 & 66.35 & 56.73 \\ \hline
\multicolumn{1}{l|}{\multirow{3}{*}{Non-motion CPLA}} & $K=2$  & 77.45 & 76.31 & 67.23 & 56.84 \\ \cline{2-6} 
\multicolumn{1}{l|}{}                                              & $K=8$ & 78.86 & 77.01 & 70.43 & 59.24 \\ \cline{2-6} 
\multicolumn{1}{l|}{}                                              & $K=16$ & 76.53 & 75.02 & 66.65 & 55.63 \\ \hline
\multicolumn{1}{l|}{\multirow{3}{*}{CPLA}}           & $K=2$  & 78.25 & 76.28 & 70.82 & 59.38 \\ \cline{2-6} 
\multicolumn{1}{l|}{}                                              & $K=8$ & \textbf{79.03} & \textbf{77.34} & \textbf{73.54} & \textbf{60.76} \\ \cline{2-6} 
\multicolumn{1}{l|}{}                                              & $K=16$ & 77.84 & 75.83 & 71.04 & 58.92 \\ \hline
\end{tabular}
\end{table}

\textbf{Modeling of spatio-temporal cues.} 
The optical flow images contain movement trend and can be leveraged for prediction along with the appearance information. We compare two different methods of modeling the fusion of two-stream cues: (1)Boost appearance detection results with motion detection scores as in \cite{suman16bmvc}, i.e. model the two-stream coupling on the bounding box level (CPLA-bbox); (2) Couple the appearance and motion information at the feature level, i.e. concatenate two stream \textit{conv5} features (CPLA-conv5). As shown in Table \ref{fusion_perf}, with explicitly modeled two-stream feature coupling, CPLA-conv5 outperforms CPLA-bbox consistently, which only uses the fusion to assist appearance detection.

\begin{table}[!htbp]
\centering
\caption{Two variants of CPLA using different modeling of two-stream cues}
\label{fusion_perf}
\begin{tabular}{l|l|l|l|l}
\hline
\multicolumn{1}{c|}{$\delta$}                           & 0.05  & 0.1   & 0.2   & 0.3   \\ \hline
CPLA-bbox & 77.33 & 74.72 & 70.39 & 58.87 \\ \hline
CPLA-conv5 & \textbf{79.03} & \textbf{77.34} & \textbf{73.54} & \textbf{60.76} \\ \hline
\end{tabular}
\end{table}

\textbf{Analysis on detection linking hyperparameter.}
The hyperparatmer $\beta$ in Equation \ref{equ:energy} affects the trade-off between detection score and spatial overlap (IoU) in the linking process. $\beta$ is set to be 0.7 empirically. Higher $\beta$ gives more weight on the relative importance of IoU and leads to more fragmented linking. Lower $\beta$ leads to ID switches in the linking. Both result in lower mAP.

\begin{figure}[!htbp]
\centering
\includegraphics[width=1.0\textwidth]{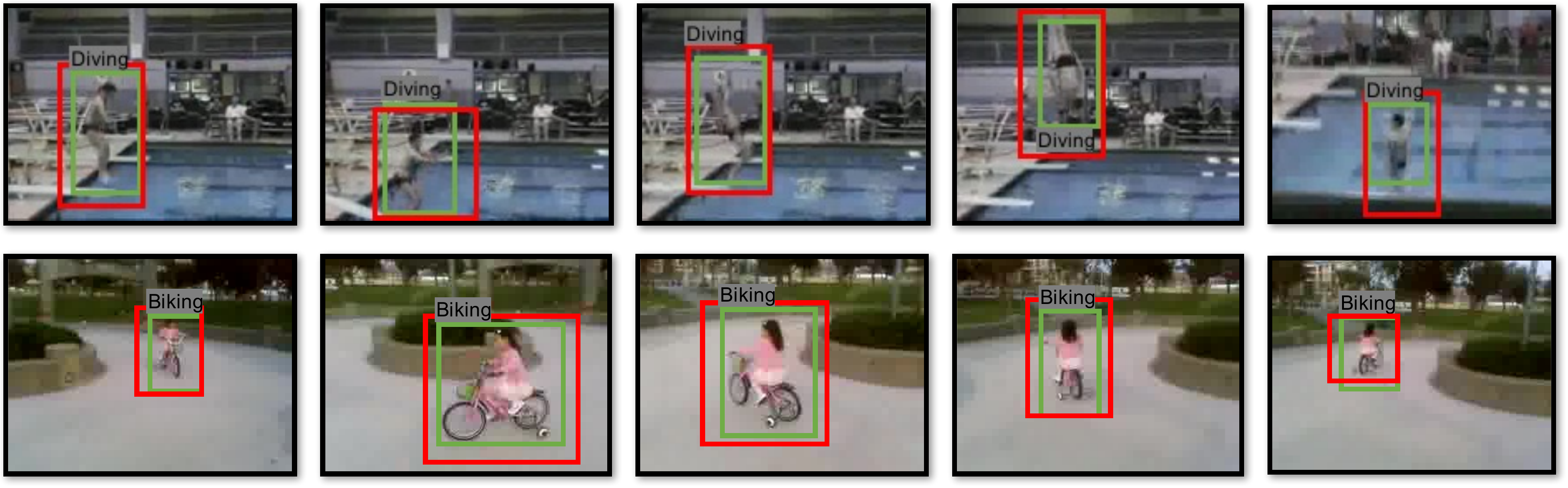}
\caption{Qualitative results on UCF101-24 video. Red bounding boxes are annotations and green bounding boxes are action detection results}
\label{fig:qualitative_results}
\end{figure}

\textbf{Comparison with state-of-the-art.} The comparisons with state-of-the-art methods on UCF101-24 and LIRIS-HARL datasets are presented in Table \ref{comp_sota} and Table \ref{liris-comp} separately. On UCF101-24, CPLA
outperforms \cite{peng2016multi} by 0.68\% at spatio-temporal IoU = 0.2. On LIRIS-HARL,  CPLA outperforms the current state-of-the-art methods by a large margin under both evaluation protocols. Some qualitative results are shown in Figure \ref{fig:qualitative_results} on UCF101-24 video.

\begin{table}[!htbp]
\centering
\caption{Quantitative action detection results on UCF101-24 dataset comparing with state-of-the-art methods.}
\label{comp_sota}
\begin{threeparttable}
\begin{tabular}{l|c|c|c|c|c|c}
\hline
\multicolumn{1}{c|}{$\delta$}    & 0.05  & 0.1   & 0.2   & 0.3   & 0.4   & 0.5   \\ \hline
\multicolumn{1}{c|}{FAP\cite{yu2015fast}}         & 42.80 & -     & -     & -     & -     & -     \\
\multicolumn{1}{c|}{STMH\cite{weinzaepfel2015learning}}        & 54.28 & 51.68 & 46.77 & 37.82 & -     & -     \\
\multicolumn{1}{c|}{Saha \textit{et al.} \cite{suman16bmvc}} & \textbf{79.12} & 76.57 & 66.75 & 55.46 & 46.35 & 35.86 \\
\multicolumn{1}{c|}{MR-TS R-CNN \cite{peng2016multi}\footnotemark} & 78.76 & 77.31 & 72.86 & \textbf{65.70} & -     & -     \\
\multicolumn{1}{c|}{CPLA}   & 79.03 & \textbf{77.34} & \textbf{73.54} & 60.76 & \textbf{49.23} & \textbf{37.80} \\ \hline
\end{tabular} 
\begin{tablenotes}\footnotesize
\item[1] Updated results of \cite{peng2016multi} from https://hal.inria.fr/hal-01349107/file/eccv16-pxj-v3.pdf
\end{tablenotes}
\end{threeparttable}
\end{table}

\begin{table}[!htbp]
\centering
\caption{Quantitative action detection results on LIRIS-HARL dataset under different metrics}
\label{liris-comp}
\fontsize{7.5}{8.5}\selectfont
\begin{tabular}{l|ccccccccc}
\hline
Methods         & Recall$_{10}$    & Precision$_{10}$ & F1-Score$_{10}$  & I$_{sr}$         & I$_{sp}$         & I$_{tr}$         & I$_{tp}$         & IQ            & mAP@$\delta=0.2$  \\ \hline
Saha \textit{et al.} \cite{suman16bmvc}     & 0.57 & 0.60          & 0.58          & 0.54          & 0.34          & 0.48          & \textbf{0.47} & 0.46          & 49.10          \\
CPLA       & \textbf{0.62}          & \textbf{0.67} & \textbf{0.70} & \textbf{0.62} & \textbf{0.42} & \textbf{0.51} & 0.44          & \textbf{0.53} & \textbf{54.34} \\ \hline
\end{tabular}
\end{table}

\section{Conclusion}

This paper introduced a cascade proposal and location anticipation (CPLA) model for spatio-temporal action detection. CPLA consists of a frame-level action detection model and a temporal linking/trimming algorithm. The action detection model takes RGB and optical flow images as input, extracts action proposals via casRPN and conducts action detection on each frame by exploiting the action region movement continuity. CPLA achieves state-of-the-art performance on both UCF101 and more challenging LIRIS-HARL datasets.

\bibliography{egbib}

\begin{thebibliography}{35}
\providecommand{\natexlab}[1]{#1}
\providecommand{\url}[1]{\texttt{#1}}
\expandafter\ifx\csname urlstyle\endcsname\relax
  \providecommand{\doi}[1]{doi: #1}\else
  \providecommand{\doi}{doi: \begingroup \urlstyle{rm}\Url}\fi

\bibitem[An et~al.(2009)An, Peursum, Liu, and Venkatesh]{an2009efficient}
Senjian An, Patrick Peursum, Wanquan Liu, and Svetha Venkatesh.
\newblock Efficient algorithms for subwindow search in object detection and
  localization.
\newblock In \emph{CVPR}, pages 264--271, 2009.

\bibitem[Bargi et~al.(2012)Bargi, Da~Xu, and Piccardi]{bargi2012online}
Ava Bargi, Richard~Yi Da~Xu, and Massimo Piccardi.
\newblock An online hdp-hmm for joint action segmentation and classification in
  motion capture data.
\newblock In \emph{CVPR Workshop}, pages 1--7. IEEE, 2012.

\bibitem[Everingham et~al.()Everingham, Van~Gool, Williams, Winn, and
  Zisserman]{pascal-voc-2007}
M.~Everingham, L.~Van~Gool, C.~K.~I. Williams, J.~Winn, and A.~Zisserman.
\newblock The {PASCAL} {V}isual {O}bject {C}lasses {C}hallenge 2007 {(VOC2007)}
  {R}esults.
\newblock
  http://www.pascal-network.org/challenges/VOC/voc2007/workshop/index.html.

\bibitem[Felzenszwalb et~al.(2008)Felzenszwalb, McAllester, and
  Ramanan]{felzenszwalb2008discriminatively}
Pedro Felzenszwalb, David McAllester, and Deva Ramanan.
\newblock A discriminatively trained, multiscale, deformable part model.
\newblock In \emph{CVPR}, pages 1--8. IEEE, 2008.

\bibitem[Forney(1973)]{forney1973viterbi}
G~David Forney.
\newblock The viterbi algorithm.
\newblock \emph{Proceedings of the IEEE}, 61\penalty0 (3):\penalty0 268--278,
  1973.

\bibitem[Gaidon et~al.(2013)Gaidon, Harchaoui, and Schmid]{gaidon2013temporal}
Adrien Gaidon, Zaid Harchaoui, and Cordelia Schmid.
\newblock Temporal localization of actions with actoms.
\newblock \emph{IEEE transactions on pattern analysis and machine
  intelligence}, 35\penalty0 (11):\penalty0 2782--2795, 2013.

\bibitem[Gan et~al.(2016)Gan, Sun, Duan, and Gong]{gan2016webly}
Chuang Gan, Chen Sun, Lixin Duan, and Boqing Gong.
\newblock Webly-supervised video recognition by mutually voting for relevant
  web images and web video frames.
\newblock In \emph{ECCV}, pages 849--866, 2016.

\bibitem[Girshick(2015)]{girshick2015fast}
Ross Girshick.
\newblock Fast r-cnn.
\newblock In \emph{ICCV}, pages 1440--1448, 2015.

\bibitem[Girshick et~al.(2014)Girshick, Donahue, Darrell, and
  Malik]{girshick14CVPR}
Ross Girshick, Jeff Donahue, Trevor Darrell, and Jitendra Malik.
\newblock Rich feature hierarchies for accurate object detection and semantic
  segmentation.
\newblock In \emph{CVPR}, 2014.

\bibitem[Gkioxari and Malik(2015)]{gkioxari2015finding}
Georgia Gkioxari and Jitendra Malik.
\newblock Finding action tubes.
\newblock In \emph{CVPR}, pages 759--768, 2015.

\bibitem[Gkioxari et~al.(2015)Gkioxari, Girshick, and
  Malik]{gkioxari2015contextual}
Georgia Gkioxari, Ross Girshick, and Jitendra Malik.
\newblock Contextual action recognition with r* cnn.
\newblock In \emph{ICCV}, pages 1080--1088, 2015.

\bibitem[He et~al.(2014)He, Zhang, Ren, and Sun]{he2014spatial}
Kaiming He, Xiangyu Zhang, Shaoqing Ren, and Jian Sun.
\newblock Spatial pyramid pooling in deep convolutional networks for visual
  recognition.
\newblock In \emph{ECCV}, pages 346--361, 2014.

\bibitem[Jain et~al.(2014)Jain, Van~Gemert, J{\'e}gou, Bouthemy, and
  Snoek]{jain2014action}
Mihir Jain, Jan Van~Gemert, Herv{\'e} J{\'e}gou, Patrick Bouthemy, and Cees~GM
  Snoek.
\newblock Action localization with tubelets from motion.
\newblock In \emph{CVPR}, pages 740--747, 2014.

\bibitem[Kingma and Ba(2015)]{kingma2014adam}
Diederik Kingma and Jimmy Ba.
\newblock Adam: A method for stochastic optimization.
\newblock In \emph{ICLR}, 2015.

\bibitem[Lu et~al.(2015)Lu, Corso, et~al.]{lu2015human}
Jiasen Lu, Jason~J Corso, et~al.
\newblock Human action segmentation with hierarchical supervoxel consistency.
\newblock In \emph{CVPR}, pages 3762--3771, 2015.

\bibitem[Ma et~al.(2016)Ma, Sigal, and Sclaroff]{ma2016learning}
Shugao Ma, Leonid Sigal, and Stan Sclaroff.
\newblock Learning activity progression in lstms for activity detection and
  early detection.
\newblock In \emph{CVPR}, pages 1942--1950, 2016.

\bibitem[Mettes et~al.(2016)Mettes, van Gemert, and Snoek]{mettes2016spot}
Pascal Mettes, Jan~C van Gemert, and Cees~GM Snoek.
\newblock Spot on: Action localization from pointly-supervised proposals.
\newblock In \emph{ECCV}, pages 437--453, 2016.

\bibitem[Peng and Schmid(2016)]{peng2016multi}
Xiaojiang Peng and Cordelia Schmid.
\newblock Multi-region two-stream r-cnn for action detection.
\newblock In \emph{ECCV}, pages 744--759, 2016.

\bibitem[Ren et~al.(2015)Ren, He, Girshick, and Sun]{ren2015faster}
Shaoqing Ren, Kaiming He, Ross Girshick, and Jian Sun.
\newblock Faster r-cnn: Towards real-time object detection with region proposal
  networks.
\newblock In \emph{NIPS}, pages 91--99, 2015.

\bibitem[Schmidt et~al.(2007)Schmidt, Fung, and Rosales]{schmidt2007fast}
Mark Schmidt, Glenn Fung, and Rmer Rosales.
\newblock Fast optimization methods for l1 regularization: A comparative study
  and two new approaches.
\newblock In \emph{ECCV}, pages 286--297, 2007.

\bibitem[Shou et~al.(2016)Shou, Wang, and Chang]{scnn_shou_wang_chang_cvpr16}
Zheng Shou, Dongang Wang, and Shih-Fu Chang.
\newblock Temporal action localization in untrimmed videos via multi-stage
  cnns.
\newblock In \emph{CVPR}, 2016.

\bibitem[Simonyan and Zisserman(2014{\natexlab{a}})]{simonyan2014two}
Karen Simonyan and Andrew Zisserman.
\newblock Two-stream convolutional networks for action recognition in videos.
\newblock In \emph{NIPS}, pages 568--576, 2014{\natexlab{a}}.

\bibitem[Simonyan and Zisserman(2014{\natexlab{b}})]{simonyan2014very}
Karen Simonyan and Andrew Zisserman.
\newblock Very deep convolutional networks for large-scale image recognition.
\newblock \emph{arXiv preprint arXiv:1409.1556}, 2014{\natexlab{b}}.

\bibitem[Singh et~al.(2016)Singh, Marks, Jones, Tuzel, and
  Shao]{singh2016multi}
Bharat Singh, Tim~K Marks, Michael Jones, Oncel Tuzel, and Ming Shao.
\newblock A multi-stream bi-directional recurrent neural network for
  fine-grained action detection.
\newblock In \emph{CVPR}, pages 1961--1970, 2016.

\bibitem[Soomro et~al.(2012)Soomro, Zamir, and Shah]{soomro2012ucf101}
Khurram Soomro, Amir~Roshan Zamir, and Mubarak Shah.
\newblock Ucf101: A dataset of 101 human actions classes from videos in the
  wild.
\newblock \emph{arXiv preprint arXiv:1212.0402}, 2012.

\bibitem[Soomro et~al.(2015)Soomro, Idrees, and Shah]{soomro2015action}
Khurram Soomro, Haroon Idrees, and Mubarak Shah.
\newblock Action localization in videos through context walk.
\newblock In \emph{CVPR}, pages 3280--3288, 2015.

\bibitem[Suman~Saha(2016)]{suman16bmvc}
Michael Sapienza Philip H. S. Torr Fabio~Cuzzlion Suman~Saha, Gurkirt~Singh.
\newblock Deep learning for detecting multiple space-time action tubes in
  videos.
\newblock \emph{BMVC}, 2016.

\bibitem[Tian et~al.(2013)Tian, Sukthankar, and Shah]{tian2013spatiotemporal}
Yicong Tian, Rahul Sukthankar, and Mubarak Shah.
\newblock Spatiotemporal deformable part models for action detection.
\newblock In \emph{CVPR}, pages 2642--2649, 2013.

\bibitem[Uijlings et~al.(2013)Uijlings, Van De~Sande, Gevers, and
  Smeulders]{uijlings2013selective}
Jasper~RR Uijlings, Koen~EA Van De~Sande, Theo Gevers, and Arnold~WM Smeulders.
\newblock Selective search for object recognition.
\newblock \emph{International journal of computer vision}, 104\penalty0
  (2):\penalty0 154--171, 2013.

\bibitem[Wang et~al.(2014)Wang, Qiao, and Tang]{wang2014video}
Limin Wang, Yu~Qiao, and Xiaoou Tang.
\newblock Video action detection with relational dynamic-poselets.
\newblock In \emph{ECCV}, pages 565--580, 2014.

\bibitem[Wang et~al.(2016)Wang, Qiao, Tang, and Van~Gool]{wang2016actionness}
Limin Wang, Yu~Qiao, Xiaoou Tang, and Luc Van~Gool.
\newblock Actionness estimation using hybrid fully convolutional networks.
\newblock In \emph{CVPR}, pages 2708--2717, 2016.

\bibitem[Weinzaepfel et~al.(2015)Weinzaepfel, Harchaoui, and
  Schmid]{weinzaepfel2015learning}
Philippe Weinzaepfel, Zaid Harchaoui, and Cordelia Schmid.
\newblock Learning to track for spatio-temporal action localization.
\newblock In \emph{CVPR}, pages 3164--3172, 2015.

\bibitem[Wolf et~al.(2014)Wolf, Lombardi, Mille, Celiktutan, Jiu, Dogan, Eren,
  Baccouche, Dellandr{\'e}a, Bichot, et~al.]{wolf2014evaluation}
Christian Wolf, Eric Lombardi, Julien Mille, Oya Celiktutan, Mingyuan Jiu, Emre
  Dogan, Gonen Eren, Moez Baccouche, Emmanuel Dellandr{\'e}a, Charles-Edmond
  Bichot, et~al.
\newblock Evaluation of video activity localizations integrating quality and
  quantity measurements.
\newblock \emph{Computer Vision and Image Understanding}, 127:\penalty0 14--30,
  2014.

\bibitem[Yu and Yuan(2015)]{yu2015fast}
Gang Yu and Junsong Yuan.
\newblock Fast action proposals for human action detection and search.
\newblock In \emph{CVPR}, pages 1302--1311, 2015.

\bibitem[Zitnick and Doll{\'a}r(2014)]{zitnick2014edge}
C~Lawrence Zitnick and Piotr Doll{\'a}r.
\newblock Edge boxes: Locating object proposals from edges.
\newblock In \emph{ECCV}, pages 391--405, 2014.

\end{thebibliography}
\end{document}